\documentclass[runningheads,a4paper]{llncs}

\usepackage{amssymb}
\usepackage{graphicx}
\usepackage{alltt}
\usepackage{wrapfig}

\usepackage{url}
\urldef{\mailsa}\path||    
\newcommand{\keywords}[1]{\par\addvspace\baselineskip
\noindent\keywordname\enspace\ignorespaces#1}

\usepackage{wrapfig}
\usepackage{url}
\usepackage{booktabs}
\usepackage{array}
\usepackage{times}   
\usepackage{multirow}

\begin{document}

\mainmatter  

\title{Narrative Variations in a Virtual Storyteller}

\titlerunning{Narrative Variations in a Virtual Storyteller}

\author{Stephanie M. Lukin
\and Marilyn A. Walker}
\authorrunning{}

\institute{Natural Language and Dialogue Systems Lab\\
University of California, Santa Cruz\\
Baskin School of Engineering\\
{\tt \{slukin,mawalker\}@ucsc.edu} 
\mailsa{}\\
}

%
%

\toctitle{Narrative Variation in a Virtual Storyteller}
\tocauthor{Lukin and Walker}
\maketitle

\begin{abstract}
Research on storytelling over the last 100 years has distinguished at
least two levels of narrative representation (1) story, or fabula; and
(2) discourse, or sujhet. We use this distinction to create {\it Fabula
  Tales}, a computational framework for a virtual storyteller that can
tell the same story in different ways through the implementation of
general narratological variations, such as varying direct vs. indirect
speech, character voice (style), point of view, and focalization. A
strength of our computational framework is that it is based on very
general methods for re-using existing story content, either from
fables or from personal narratives collected from blogs.  We first
explain how a simple annotation tool allows na\'ive annotators to easily
create a deep representation of fabula called a story intention graph,
and show how we use this representation to generate story tellings
automatically.  Then we present results of two studies testing our
narratological parameters, and showing that different tellings affect
the reader's perception of the story and characters.

\end{abstract}

\section{Introduction}

Research on oral storytelling over the last 100 years has
distinguished at least two levels of narrative representation (1)
story, or fabula: the content of a narrative in terms of the sequence
of events and relations between them, the story characters and their
traits and affects, and the properties and settings; and (2) discourse,
or sujhet: the actual expressive telling of a story as a stream of
words, gestures, images or facial expressions in a storytelling medium
\cite{Bal81,Shklovsky91,Genette83,Prince73,Propp68}. In the telling of
a narrative, events from the story are selected, ordered, and
expressed in the discourse.  We use this distinction to create {\it
  Fabula Tales}, a computational framework for a virtual storyteller
that can tell the same story in different ways, using a set of general
narratological variations, such as direct vs. indirect speech,
character voice (style), point of view, and focalization.

We demonstrate the generality of our methods by applying them to both
Aesop's Fables and personal narratives from a pre-existing corpus of
blogs \cite{GordonSwanson09}. We hypothesize many advantages for a
virtual storyteller who can repurpose existing stories. Stories such as {\it
  The Startled Squirrel} in Fig.~\ref{squirrel-blog-story} are created daily in the thousands and cover any topic
imaginable. They are natural and personal, and may be funny, sad,
heart-warming or serious. 
Applications for virtual
storytellers who can retell these stories in different ways could include 
virtual companions, persuasion, educational storytelling, or sharing
troubles in therapeutic settings
\cite{Bickmore03,Traumetal07,PennebakerSeagal99,Gratchetal12,SlaterRouner02}.
Fig.~\ref{nv-v1-fig} shows how {\it Fabula Tales} can shift from third person to first person 
automatically using content from {\it The Startled Squirrel}
(Fig.\ref{squirrel-blog-story}). To our knowledge, this is the first
time that these narratological variations have been implemented in a
framework where the discourse (telling) is completely independent of
the fabula (content) of the story \cite{loenneker2005narratological}.

\begin{figure}[t!h]
\centering
\begin{small}
\vspace{-0.2in}
\begin{tabular}{|p{4.65in}|}
\hline 
This is one of those times I wish I had a digital camera. We keep a
large stainless steel bowl of water outside on the back deck for
Benjamin to drink out of when he's playing outside. His bowl has
become a very popular site. Throughout the day, many birds drink out
of it and bathe in it. The birds literally line up on the railing
and wait their turn. Squirrels also come to drink out of it. The
craziest squirrel just came by- he was literally jumping in fright at
what I believe was his own reflection in the bowl. He was startled so
much at one point that he leap in the air and fell off the deck. But
not quite, I saw his one little paw hanging on! After a moment or
two his paw slipped and he tumbled down a few feet. But oh, if you
could have seen the look on his startled face and how he jumped back
each time he caught his reflection in the bowl!  \\
\hline
\end{tabular}
\vspace{-0.1in}
\caption{{\it The Startled Squirrel} personal narrative \label{squirrel-blog-story}}
\vspace{-0.3in}
\end{small}
\end{figure}

\begin{figure}[th!]
\centering
\begin{small}
\vspace{-0.2in}
\begin{tabular}{|c|p{4.5in}|}
\hline 
\bf ID &  \bf Example \\
\hline \hline 
\bf S1 &  The narrator placed the bowl on the deck in order for
Benjamin to drink the bowl's water. The bowl was popular. The birds
drank the bowl's water. The birds bathed themselves in the bowl. The
birds organized themselves on the deck's railing in order for the
birds to wait. \\ \hline
\bf S2 & I approached the bowl. I was startled because I saw
my reflection. I leaped because I was startled. I fell over the
deck's railing because I leaped because I was startled. I held the
deck's railing with my paw. My paw slipped off the deck's
railing. I fell. \\ \hline
\end{tabular}
\vspace{-0.1in}
\caption{Variation in Point of View for {\it The Startled Squirrel} \label{nv-v1-fig}}
\end{small}
\vspace{-0.2in}
\end{figure}

%

Sec.~\ref{sig-sec} describes how the deep structure of any narrative
can be represented as a story intention graph, a generic model of the fabula
\cite{elsonthesis}. Sec.~\ref{method-sec} describes our method for
generating retellings of stories, and Sec.~\ref{res-sec}
describes two experimental evaluations.  We delay discussion of
related work to Sec.~\ref{conc-sec} when we can compare it to our own,
and sum up and discuss future work.

\section{Repurposing Stories with Story Intention Graphs}
\label{sig-sec}

\begin{figure}
\vspace{-.3in}
\begin{small}
\begin{tabular}{|p{4.75in}|}
\hline
A Crow was sitting on a branch of a tree with a piece of cheese in her beak when a Fox observed her and set his wits to work to discover some way of getting the cheese.
Coming and standing under the tree he looked up and said, ``What a noble bird I see above me! Her beauty is without equal, the hue of her plumage exquisite. If only her voice is as sweet as her looks are fair, she ought without doubt to be Queen of the Birds.''
The Crow was hugely flattered by this, and just to show the Fox that she could sing she gave a loud caw.
Down came the cheese,of course, and the Fox, snatching it up, said, ``You have a voice, madam, I see: what you want is wits.'' \\ \hline
\end{tabular}
\vspace{-.175in}
\caption{\label{fc-original} The Fox and The Crow}
\end{small}
\vspace{-0.2in}
\end{figure}

\begin{wrapfigure}{r}{0.7\textwidth}
\vspace{-0.1in}
\centering
\includegraphics[width=3.5in]{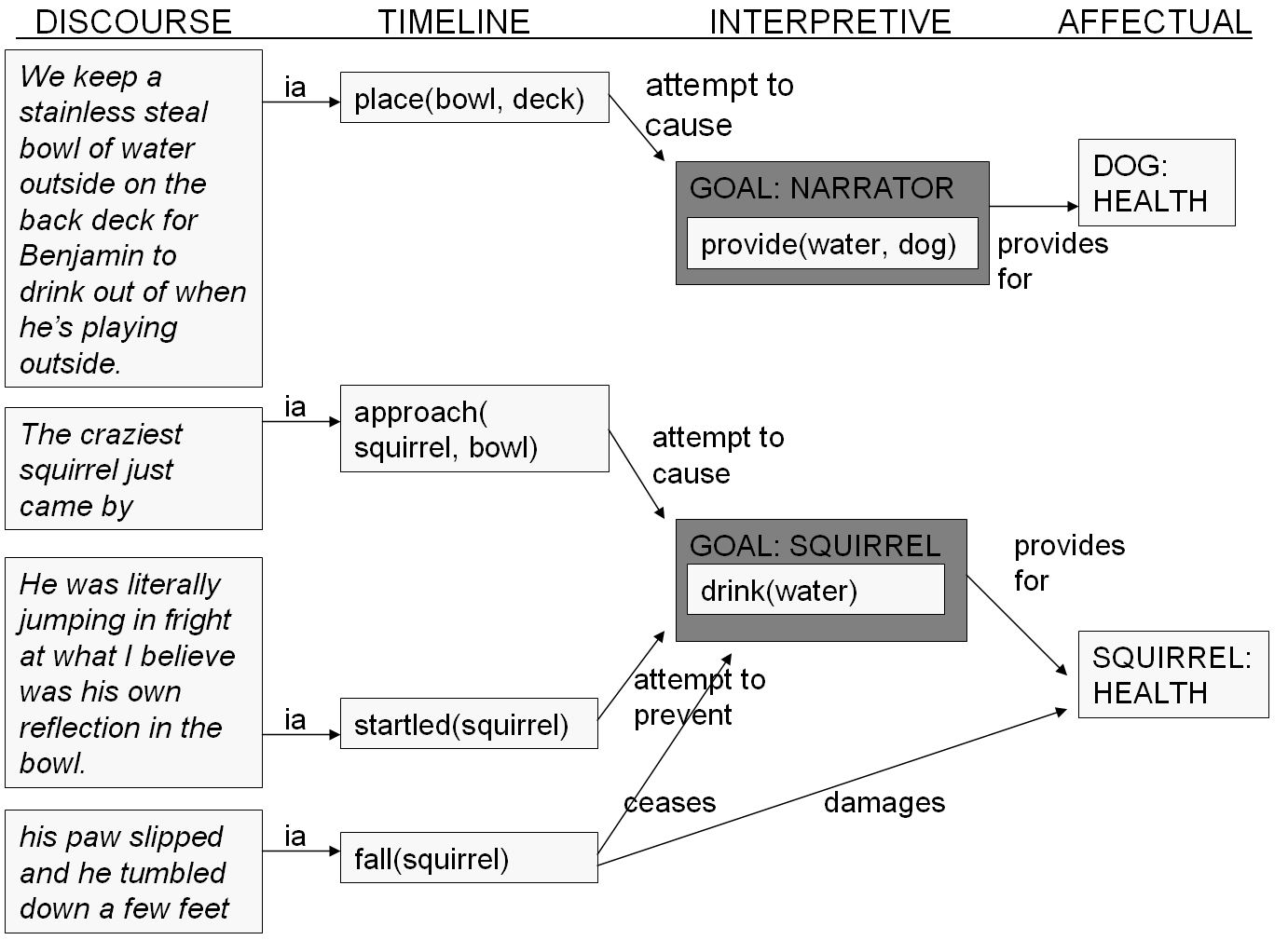}
\vspace{-0.215in}
\caption{\label{squirrel-sig} Part of the {\sc Story Intention Graph} ({\sc sig}) for {\it The Startled Squirrel}.}
\vspace{-0.35in}
\end{wrapfigure}

Our framework builds on Elson's representation of fabula, called a
story intention graph, or {\sc sig} \cite{elsonthesis}. 
The {\sc sig} allows many aspects of a story to be captured, including key
entities, events and statives arranged in a timeline, and an interpretation of the
overarching goals, plans and beliefs of the story's agents \cite{elsonthesis}. 
Fig.~\ref{squirrel-sig} shows the part of the {\sc sig} for {\it The Startled
  Squirrel} story in Fig.~\ref{squirrel-blog-story}.  
Elson's {\sc dramabank} provides 36 Aesop's Fables encoded as {\sc sig}s,
e.g. {\it The Fox and the Crow} in Fig.~\ref{fc-original}, and Elson's 
annotation tool Scheherazade allows minimally trained annotators to develop a {\sc sig} for any
narrative. We hired an undergraduate linguist to use Scheherezade to
produce {\sc sig}s for 100 personal narratives. Each story took on
average 45 minutes to annotate.  We currently have 100 annotated
stories on topics such as travel, daily activities, storms, gardening,
funerals, going to the doctor, camping, and snorkeling.


Scheherazade allows users to annotate a story
along several dimensions, starting with the surface form, or discourse
as shown in Fig.~\ref{squirrel-sig}, and then proceeding to
deeper representations.  The second column in
Fig.~\ref{squirrel-sig} is called the ``timeline layer'', in which
the story facts are encoded as predicate-argument structures
(propositions) and temporally ordered on a timeline. The timeline
layer consists of a network of propositional structures, where nodes
correspond to lexical items that are linked by thematic
relations. Scheherazade adapts information about predicate-argument
structures from the VerbNet lexical database
\cite{kipper2006extensive} and uses WordNet \cite{fellbaum2010wordnet}
as its noun and adjectives taxonomy. The arcs of the story graph are
labeled with discourse relations. Scheherazade also comes with a built-in realizer (referred to as {\it sch} in this paper) that the annotator can use to check their work. This realizer does not incorporate any narratological variations.


\section{Generating Narratological Variations}
\label{method-sec}

Our framework can generate story re-tellings using methods that are
neither genre nor domain-specific. We build {\it Fabula Tales} on two tools from previous
work: {\sc personage} and the ES-Translator \cite{MairesseWalker11,rishes2013generating}. 
{\sc personage} is an expressive
natural language generation engine that takes as input the syntactic 
formalism of Deep Syntactic Structures ({\sc dsynts}) \cite{lavoie1997fast,igorʹ1988dependency}.
{\sc dsynts} allow {\sc personage} to be flexible in generation, however the 
creation of {\sc dsynts} has been hand crafted and time consuming.
The ES-Translator ({\sc est}) automatically bridges the
narrative representation of the {\sc sig} to the {\sc dsynts} formalism by applying a model of syntax to the {\sc sig} \cite{rishes2013generating}. The {\sc sig} representation
gives us direct access to the linguistic and logical representations
of the fabula for each story, so the {\sc est} can interpret the story in the {\sc dsynts} formalism and 
retell it using different words or syntactic structures
~\cite{rishes2013generating,lukin2014automating}.

\begin{figure}[ht]
\centering
\includegraphics[width=4.0in]{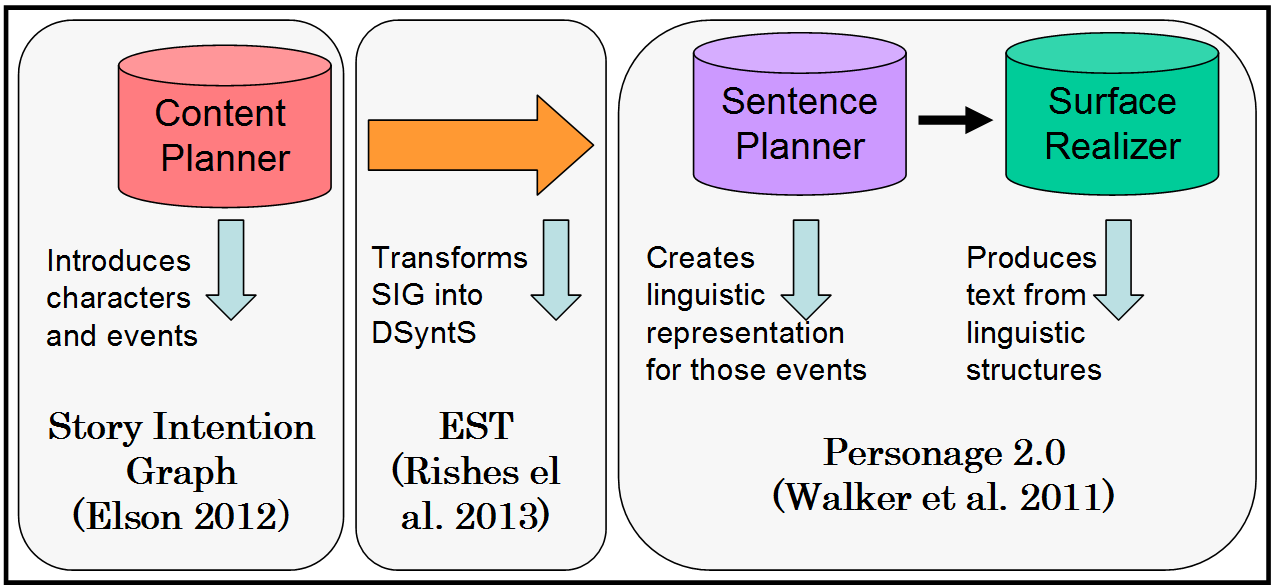}
\vspace{-0.1in}
\caption{\label{est-arch-fig} {NLG pipeline method of the ES Translator.}}
\vspace{-0.2in}
\end{figure}

{\sc dsynts} are dependency structures where the nodes are labeled
with lexemes and the arcs of the tree are labeled with syntactic
relations. The {\sc dsynts} formalism distinguishes between arguments and
modifiers and between argument types (subject, direct
and indirect object etc). {\sc personage} handles
morphology, agreement and function words to produce an output
string.

After the {\sc est} applies syntax to the {\sc sig}, it generates two
data structures: text plans containing sentence plans and the
corresponding {\sc dsynts}. Thus any story or content represented as
a {\sc sig} can be retold using {\sc personage}.
Fig.~\ref{est-arch-fig} provides a high level view of the architecture
of {\sc est}. The full translation methodology is described in
\cite{rishes2013generating}. 

This paper incorporates the {\sc est} pipeline (including {\sc sig}s and {\sc personage})
into the {\it Fabula Tales} computational framework and adds
three narratological parameters into story generation:
\begin{enumerate}
\item {\bf Point of View:} change the narration point of view to any character in a story in the first person voice (Sec.~\ref{pov-sec}.)
\item {\bf Direct Speech:} given any {\sc sig} encoding that uses
  speech act verbs (e.g. said, told, asked, alleged), re-tell
as  direct speech or indirect speech (Sec.~\ref{ds-sec}.)
\item {\bf Character Voice:} Substitute different character voices using
any character model expressible with {\sc personage}'s 67 parameters (Sec.~\ref{cv-sec}.)
\end{enumerate}

Fig.~\ref{nv-v2-fig} provides variations that combine these
narratological parameters illustrating content from ``The Fox and the Crow'' and two additional
stories: Conflict at Work, and The Embarrassed Teacher. B2 and C1 are examples of the original tellings and C2 is a {\it sch} realization.

\begin{figure}[th!]
\centering
\begin{small}
\begin{tabular}{|p{.75in}|c|p{.95in}|p{2.75in}|}
\hline 
\bf Narr Param & \bf ID &  \bf Content & \bf Example \\
\hline \hline 
Direct Speech & A1 &  Fox and Crow & 
The crow sat on the tree's branch. The cheese was in the crow's pecker. The crow thought ``I will eat the cheese on the branch of the tree because the clarity of the sky is so-somewhat beautiful."\\ \hline
Direct Speech & B1 & Conflict at Work &  ``The company requires the division to sign the document", the director told the division. ``Be expedient", the director told the division.  \\ \hline
Original & B2 & Conflict at Work & 
  The new director sent out an email noting the urgency of everyone
  signing, scanning, and formatting the signed and scanned contract
  into a PDF. He noted that it had to be done that very day (a
  Friday). \\ \hline
Original & C1 & Embarrassed Teacher & 
I had taken the register and was standing at the front of the
  class doing some revision... However, all eyes were not on my face
  but at my ankles. Nervously I looked down to see that my underslip
  had somehow made its way to the floor. Elastic gone What to do?. \\ \hline 
Sch & C2 & Embarrassed Teacher & The narrator lifted the slip and inserted it into a bottom drawer of the desk. The narrator resumed teaching, and the group of students didn't react.   \\ \hline
Indirect Speech & A2 & Fox and the Crow &  The fox said the beauty of the bird was incomparable. The fox said the hue of the feather of the bird was exquisite. 
\\ \hline
Indirect Speech & B3 & Conflict at Work & 
The narrator said if the director said the thing was urgent the narrator would need to be urgent. The narrator said the director was frivolous.  \\ \hline
Character Voice & A3 &  Fox and the Crow & 
The fox alleged ``your beauty is quite incomparable, okay?" The fox alleged ``your feather's chromaticity is damn exquisite." 
\\ \hline
Character Voice & C3 & Embarrassed Teacher & I stood at the classroom's front. I no-noticed my ankle to be somewhat observed. I looked nervously toward my ankle. I glanced around the students. 
\\ \hline 
Point of View &  A4 &  Fox and the Crow & 
I sat on the tree's branch. The cheese was in my beak. The fox observed me. 
The fox came. The fox stood under the tree. The fox looked toward me. The fox said he saw me. \\ \hline
\end{tabular}
\end{small}
\caption{Narratological Variations in Blogs and Aesops \label{nv-v2-fig}}
\vspace{-0.1in}
\end{figure}

\subsection{Point of View}
\label{pov-sec}

From the deep syntactic structure in the
format of {\sc dsynts}, we can change the narration style from
the third person perspective to the first person perspective of any character
in the story (see example A4 in Fig.~\ref{nv-v2-fig}). We define simple rules to
make this transformation within the {\sc dsynts} itself, not at the
sentence level. Table~\ref{dsynts} shows the {\sc dsynts},
which are represented as xml structures, for the sentence {\it The
  crow flew herself to the window}.

In order to transform the sentence into the
first person, only simple changes to the deep structure are necessary. At
lines 9 and 10 in Table~\ref{dsynts}, we assign the {\tt person} attribute to {\tt 1st} to
specify a change of point of view to first person. The surface realizer in {\sc personage} takes care
of the transformations with its own rules, knowing to change whatever
lexeme is present at line 9 simply to {\it I}, and to change the
coreference resolutions at line 10 to {\it myself}. This is a major
advantage of our computational framework: the deep linguistic representation allows us
to specify changes we want without manipulating strings, and allows
general rules for narratological parameters such as voice.

\subsection{Dialogue Realization}
\label{ds-sec}

\begin{wrapfigure}{r}{1.75in}
\vspace{-0.2in}
\centering
\includegraphics[width=1.75in]{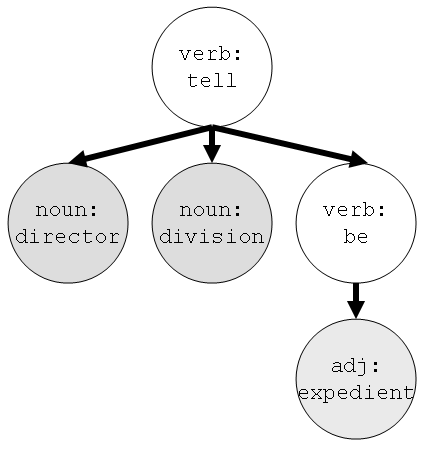}
\caption{{\it The director told the division to be expedient.} \label{ds1}}
\includegraphics[width=1.75in]{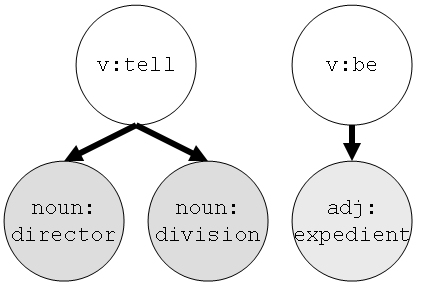}
\caption{{\it ``Be expedient", the director told the division} or {\it The director told the division``be expedient."} \label{ds2}}
\vspace{-0.35in}
\end{wrapfigure}

By default, speech acts in the {\sc sig} are encoded as indirect
speech. We automatically detect a speech act from its verb type in the
WordNet online dictionary, and then transform it to a direct speech
act (see A1, A2, B1, and B3 in Fig.~\ref{nv-v2-fig}).  First we use
WordNet to identify if the main verb in a sentence is a verb of
communication.  Next, we break apart the {\sc dsynts} into their tree
structure (Fig.~\ref{ds1}). For example, we first identify the subject
({\it director}) from utterance B1 in Fig.~\ref{nv-v2-fig}, and object
({\it division}) of the main verb of communication ({\it tell}). Then
we identify the remainder of the tree ({\it be} is the root verb), which is
what is to be uttered, and split it off from its parent verb of
communication node, thus creating two separate {\sc dsynts}
(Fig.~\ref{ds2}). In {\sc personage}, we create a direct speech text
plan to realize the explanatory in the default narrator style and the
utterance in a specified character voice and appropriately insert the
quotation marks. We can then realize direct speech as {\it ``Utterance" said X.} or {\it X said ``utterance."}

\begin{table}[h!]
\centering
\begin{scriptsize}
\caption{{\sc dsynts} for ``The crow flew herself to the window'' and ``I flew myself to the window'' \label{dsynts}}
\vspace{-0.1in}
\begin{tabular}{|p{4.8in}|}
\hline 
``The crow flew herself to the window'' 
\vspace{-0.1in}
\begingroup
\fontsize{8pt}{8pt}
\begin{verbatim}
1 <dsyntnode class="verb" lexeme="fly">
2   <dsyntnode class="common_noun" lexeme="crow" gender="fem">
3   <dsyntnode class="common_noun" lexeme="crow" gender="fem" 
               pro="pro">
4   <dsyntnode class="preposition" lexeme="to">
5     <dsyntnode class="common_noun" lexeme="window">
6   </dsyntnode>
7 </dsyntnode>
\end{verbatim} 
\endgroup
\\ \hline
``I flew myself to the window'' 
\begingroup
\fontsize{8pt}{8pt}
\begin{alltt}
8 <dsyntnode class="verb" lexeme="fly">
9     <dsyntnode class="common_noun" lexeme="crow" gender="fem" 
      \emph{person="1st"}>
10   <dsyntnode class="common_noun" lexeme="crow" gender="fem" 
      pro="pro" \emph{person="1st"}>
11   <dsyntnode class="preposition" lexeme="to">
12     <dsyntnode class="common_noun" lexeme="window">
13  </dsyntnode>
14</dsyntnode>
\end{alltt} 
\endgroup
\\ \hline
\end{tabular}
\end{scriptsize}
\vspace{-0.1in}
\end{table}

\vspace{0.5in}
\subsection{Character Voice}
\label{cv-sec}

The main advantage of {\sc personage} is its ability to generate a
single utterance in many different voices.  Models of narrative style
are currently based on the Big Five personality traits
\cite{MairesseWalker11}, or are learned from film scripts
\cite{Walkeretal11}. Each type of model (personality trait or film)
specifies a set of language cues, one of 67 different parameters,
whose value varies with the personality or style to be
conveyed. Previous work in \cite{MairesseWalker11} has shown that humans
perceive the personality stylistic models in the way that {\sc
  personage} intended, and \cite{Walkeretal11} shows that character
utterances in a new domain can be recognized by humans
as models based on a particular film character.

After we add new rules to {\it Fabula Tales} to handle direct speech, we modified the original {\sc sig} representation of
the {\it Fox and the Crow} to contain more dialogue in order to
evaluate a broader range of character styles, along with the use of direct speech.
Table~\ref{pers-fig} shows a subset of parameters,
which were used in the three personality models we tested here: the
{\it laid-back} model for the fox's direct speech, the {\it shy} model
for the crow's direct speech, and the {\it neutral} model for the
narrator voice. The {\it laid-back} model uses emphasizers, hedges,
exclamations, and expletives, whereas the {\it shy} model uses
softener hedges, stuttering, and filled pauses. The {\it neutral}
model is the simplest model that does not utilize any of the extremes
of the {\sc personage} parameters.

\begin{table}[!htb]
\begin{scriptsize}
\centering \caption{\label{pers-fig} Examples of pragmatic marker
insertion parameters from {\sc personage}}
\vspace{-0.1in}
\begin{tabular}{|p{0.55in}|p{0.85in}p{1.83in}|p{1.45in}|}
\hline
{\bf Model} & {\bf Parameter} & {\bf Description} & {\bf Example}\\
\hline
\multirow{3}{*}{Shy} & {\sc Softener hedges} & Insert syntactic elements ({\it sort of},
{\it  kind of}, {\it somewhat}, {\it quite}, {\it around}, {\it
rather}, {\it I think that}, {\it it seems that}, {\it it seems to
me that}) to mitigate the strength of a proposition & {\it `It seems to me that he was hungry'}\\ 
&{\sc Stuttering} &  Duplicate parts of a content word & {\it `The vine hung on the tr-trellis'}\\
&{\sc Filled pauses} & Insert  syntactic elements expressing
hesitancy ({\it I  mean},  {\it err},  {\it mmhm}, {\it like}, {\it
you~know}) & {\it `Err... the fox jumped'}\\
\hline
\multirow{3}{*}{Laid Back} & {\sc Emphasizer hedges} & Insert  syntactic elements ({\it really},
{\it basically}, {\it actually}) to strengthen a
proposition & {\it `The fox failed to get the group of grapes, alright?'}  \\ 
& {\sc Exclamation} & Insert an exclamation mark & {\it `The group of grapes hung on the vine!'} \\
& {\sc Expletives} & Insert  a swear word & {\it `The fox was damn hungry'}\\
\hline
\end{tabular}
\end{scriptsize}
\vspace{-0.2in}
\end{table}

C3 in Fig.~\ref{nv-v2-fig} provides an example of {\it Fabula Tales} rendering a story in a
single voice for {\it The Embarrassed Teacher}. We tell the story from her
point of view and give her an introverted voice. We also show that we
can specify voices for characters in dialogue as in the Fable
excerpt in A3 in Fig.~\ref{nv-v2-fig}. {\it Fabula Tales} system allows multiple
personalities to be loaded and assigned to characters so that {\sc
  personage} runs once, fully automatically, and {\bf in real-time}.

\section{Experimental Results}
\label{res-sec}

We present two experiments that show how the flexibility of the {\sc est} combined with our narratological parameters to create {\it Fabula Tales} allows us to manipulate the perception of characters and story engagement and interest. We first present {\it The Fox and the Crow} with variations on direct speech and voice, followed by {\it Embarrassed Teacher} with variations on voice and point of view.

\subsection{Perceptions of Voice and Direct Speech}

We collect user perceptions of the {\it The Fox and the Crow}
generated with direct speech and with different personality models
(character voices) for each speech act. A dialogic variation plus
character voice excerpt is A3 in Fig.~\ref{nv-v2-fig}.  The dialogic
story is told 1) only with the neutral model; 2) with the crow
as shy and the fox as laid-back; and 3) with the crow
as laid-back and the fox as shy.

\begin{wraptable}{r}{2.5in}
\vspace{-0.5in}
\centering
\begin{small}
\caption{\label{crow-eval} Polarity of Adjectives describing the Crow and Fox (\% of total words)}
\begin{tabular}{|r|c|c||r|c|c|} \hline
Crow  & Pos & Neg & Fox & Pos & Neg \\ \hline
Neutral & 13 & 29 & Neutral & 38 & 4 \\
Shy & 28 & 24 & Shy & 39 & 8 \\
Laid-back & 10 & 22 & Laid-back & 34 & 8 \\ \hline
\end{tabular}
\end{small}
\vspace{-0.2in}
\end{wraptable}

Subjects are given a free text box and asked to enter as many
words as they wish to use to describe the characters in the story.
Table~\ref{crow-eval} shows the percentage of positive and negative
descriptive words when categorized by LIWC
\cite{pennebaker2001linguistic}. Some words include ``clever'' and ``sneaky'' for the laid-back and neutral fox, and ``shy'' and ``wise'' for the shy fox. The laid-back and neutral crow was pereived as ``na\'ive'' and ``gullible'' whereas the shy crow is more ``stupid'' and ``foolish''.

Overall, the crow's shy voice is perceived as more positive
than the crow's neutral voice, (ttest(12) = -4.38, p $<$ 0.0001), 
and the crow's laid-back voice
(ttest(12) = -6.32, p $<$ 0.0001). We hypothesize that this is because the stuttering and hesitations make the
character seem more helpless and tricked, rather than the
  laid-back model which is more boisterous. However, there
is less variation between the fox polarity. Both the
stuttering shy fox and the boisterous laid-back fox were
seen equally as ``cunning" and ``smart".
Although we don't observe a difference between all characters, there is enough evidence to 
warrent further investigation of how reader perceptions change when the same 
content is realized in difference voices.

\subsection{Perceptions of Voice and POV}

In this experiment, we aim to see how different points of view and
voices effect reader engagement and interest. We present readers with
a one sentence summary of the {\it Embarrassed Teacher} story and 6
retellings of a sentence from the story, framed as ``possible excerpts
that could come from this summary''. We show retellings of a sentence
from {\it Embarrassed Teacher} in first person neutral, first person
shy, first person laid-back, third person neutral, the original story,
and {\it sch}. We ask participants to rate each excerpt for their
interest in wanting to read more of the story based on the style and
information given in the excerpt, and to indicate their engagement
with the story given the excerpt.
\begin{wrapfigure}{r}{3.5in}
\vspace{-0.2in}
\begin{center}
\begin{tabular}{|r|c|c|c|c|c|c|}
\hline
Engagement & Orig & 1st-out & 1st-neutr & 1st-shy & sch & 3rd-neutr \\ \hline \hline
M & 3.98 & 3.27 & 3.00 & 2.73 & 1.95 & 1.93 \\ \hline
SD & 1.07 & 1.39 & 1.19 & 1.25 & 1.07 & 1.06 \\ \hline 
\multicolumn{7}{c}{} \\ \hline
Interest & Orig & 1st-out & 1st-neutr & 1st-shy & sch & 3rd-neutr \\ \hline \hline
Mean & 3.91 & 3.02 & 3.02 & 2.81 & 1.90 & 1.87 \\ \hline
SD & 0.99 & 1.21 & 1.37 & 1.27 & 1.05 & 1.01 \\ \hline
\end{tabular}
\caption{Means (M) and standard deviation (SD) for engagement and interest for original sentences and all variations in Perceptions of Voice and POV Experiment
\label{exp2-stats} }
\end{center}
\vspace{-0.1in}
\end{wrapfigure}

\begin{wrapfigure}{r}{2.75in}
\vspace{-0.45in}
\begin{center}
\includegraphics[width=2.75in]{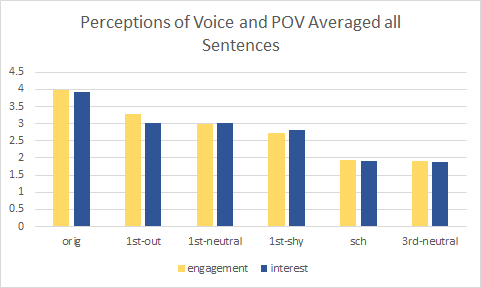}
\vspace{-0.3in}
\caption{\label{hist1} Histogram of Engagement and Interest for Perceptions of Voice and POV Experiment averaged across story (higher is better)}
\end{center}
\vspace{-0.4in}
\end{wrapfigure}

Fig.~\ref{exp2-stats} shows the means and standard deviation for engagement and interest ratings. We find a clear ranking for engagement: the original sentence is scored highest, followed by first outgoing, first neutral, first shy, {\it sch}, and third neutral. 

Fig.~\ref{hist1} shows the average engagement and interest for all the sentences. For engagement, paired t-tests show that there is a significant difference between original and first outgoing (ttest(94) = -3.99, p $<$ 0.0001), first outgoing and first shy (ttest(94) = 3.71, p $<$ 0.0001), and first shy and {\it sch} (ttest(94) = 5.60, p $<$ 0.0001). However, there are no differences between first neutral and first outgoing (ttest(95) = -1.63, p $=$ 0.05), and {\it sch} and third neutral (ttest(94) = -0.31, p $=$ 0.38). We also performed an ANOVA and found there is a significant effect on style (F(1) = 224.24, p $=$ 0), sentence (F(9) = 5.49, p $=$ 0), and an interaction between style and sentence (F(9) =1.65, p $<$ 0.1). 

For interest, we find the same ranking: the original sentence, first outgoing, first neutral, first shy, {\it sch}, and third neutral. Paired t-tests for interest show a significant difference between original and first outgoing (ttest(93) = 5.59, p $<$ 0.0001), and first shy and {\it sch} (ttest(93) = 6.16, p $<$ 0.0001). There is no difference between first outgoing and first neutral (ttest(93) = 0, p $<$ 0.5), first neutral and first shy (ttest(93) = 2.20, p $=$ 0.01), and {\it sch} and third neutral (ttest(93) = 0.54, p $=$ 0.29). We also performed an ANOVA and found there is a significant effect on style (F(1) = 204.08, p $=$ 0), sentence (F(9) = 7.32, p $=$ 0), and no interaction between style and sentence (F(9) =0.64, p $=$ 1).

We also find qualitative evidence that there are significant
differences in reader's interest and engagement in a story dependent
only upon the style. Readers preferred to read this story in the first
person: ``[the] immediacy of first person ... excerpts made me feel I
was there'', ``I felt as though those that had more detail and were
from a personal perspective were more engaging and thought evoking
versus saying the narrator did it'', and ``I felt more engaged and
interested when I felt like the narrator was speaking to me directly,
as I found it easier to imagine the situation.'' This further supports
our hypothesis that our framework to change POV will effect reader
perceptions.

Readers also identified differences in the style of the voice. Two readers commented about first outgoing: ``The `oh I resumed...' Feels more personal and is more engaging'' and  ``curse words are used to express the severity of the situation wisely''. About first shy, ``Adding the feeling of nervousness and where she looked made sense''. This suggests that certain styles of narration are more appropriate or preferred than others given the context of the story.

\vspace{-0.2in}
\section{Discussion and Future Work}
\label{conc-sec}
We introduce {\it Fabula Tales}, a computational framework for story
generation that produces narratological variations of the same story
from the fabula. We present examples showing that the capability we
have developed is general, and can be applied to informal personal
narratives.  We present experiments showing that these novel
narratological parameters lead to different perceptions of the story.
Our approach builds on previous work which focused on generating
variations of Aesop's Fables such as {\it The Fox and the Crow}
\cite{rishes2013generating}, however this previous work did not carry
out perceptual studies.

Previous work has dubbed the challenges of generating different story
tellings from fabula the {\bf NLG gap}: an architectural disconnect
between narrative generation (fabula) and natural language generation
(sujet) \cite{loenneker2005narratological,callaway2002narrative}.  To
our knowledge, there are only two previous lines of research that
address the NLG gap.  The {\sc storybook} generator is an end-to-end
narrative prose generation system that utilizes a primitive narrative
planner along with a generation engine to produce stories in the Little
Red Riding Hood fairy tale domain \cite{callaway2002narrative}. This
work manipulates NLG parameters such as lexical choice and syntactic
structure, as well as narratological parameters such as person and
focalization and the choice of whether to realize dialogue as direct
or indirect speech.  Similarly the IF system can generate multiple
variations of text in an interactive fiction (IF) environment
\cite{montfort2007generating}.  The IF system (and its successor
Curveship) uses a world simulator as the fabula, and renders narrative
variations, such as different focalizations or temporal orders.
However {\sc storybook} can only generate stories in the domain of
Little Red Riding Hood, and IF can only generate stories in its
interactive fiction world.  Other work implements narratological
variations in the story planner and does not attempt to bridge the NLG
gap \cite{bae2011toward}.  

In future work, we aim to further develop {\it Fabula Tales} and to
test in more detail the perceptual effects of narratological
variations on user interpretations of a story. Furthermore, we hope to learn
when certain styles are preferred given the context in the {\sc sig}.

\vspace{.1in}
\noindent{\bf Acknowledgments} This research was supported by NSF Creative IT program grant \#IIS-1002921, and a grant from the Nuance Foundation.

\bibliographystyle{plain}

\begin{thebibliography}{10}

\bibitem{bae2011toward}
Byung-Chull Bae, Yun-Gyung Cheong, and R~Michael Young.
\newblock Toward a computational model of focalization in narrative.
\newblock In {\em Proceedings of the 6th International Conference on
  Foundations of Digital Games}, pages 313--315. ACM, 2011.

\bibitem{Bal81}
Mieke Bal and Eve Tavor.
\newblock Notes on narrative embedding.
\newblock {\em Poetics Today}, pages 41--59, 1981.

\bibitem{Bickmore03}
T.W. Bickmore.
\newblock {\em {Relational agents: Effecting change through human-computer
  relationships}}.
\newblock PhD thesis, MIT Media Lab, 2003.

\bibitem{callaway2002narrative}
Charles~B Callaway and James~C Lester.
\newblock Narrative prose generation.
\newblock {\em Artificial Intelligence}, 139(2):213--252, 2002.

\bibitem{elsonthesis}
David Elson.
\newblock {\em Modeling Narrative Discourse}.
\newblock PhD thesis, 2012.

\bibitem{fellbaum2010wordnet}
Christiane Fellbaum.
\newblock Wordnet: An electronic lexical database. 1998.
\newblock {\em WordNet is available from http://www. cogsci. princeton.
  edu/wn}, 2010.

\bibitem{Genette83}
G{\'e}rard Genette.
\newblock {\em Nouveau discours du r{\'e}cit}.
\newblock {\'E}d. du Seuil, 1983.

\bibitem{GordonSwanson09}
Andrew Gordon and Reid Swanson.
\newblock Identifying personal stories in millions of weblog entries.
\newblock In {\em Third International Conference on Weblogs and Social Media,
  Data Challenge Workshop, San Jose, CA}, 2009.

\bibitem{Gratchetal12}
Jonathan Gratch, Louis-Philippe Morency, Stefan Scherer, Giota Stratou, Jill
  Boberg, Sebastian Koenig, Todd Adamson, Albert Rizzo, et~al.
\newblock User-state sensing for virtual health agents and telehealth
  applications.
\newblock {\em Studies in health technology and informatics}, 184:151--157,
  2012.

\bibitem{igorʹ1988dependency}
Aleksandrovi{\v{c}}~Mel'{\v{c}}uk Igor'.
\newblock {\em Dependency syntax: theory and practice}.
\newblock SUNY Press, 1988.

\bibitem{kipper2006extensive}
Karin Kipper, Anna Korhonen, Neville Ryant, and Martha Palmer.
\newblock Extensive classifications of english verbs.
\newblock In {\em Proceedings of the 12th EURALEX International Congress},
  pages 1--15, 2006.

\bibitem{lavoie1997fast}
Benoit Lavoie and Owen Rambow.
\newblock A fast and portable realizer for text generation systems.
\newblock In {\em Proceedings of the fifth conference on Applied natural
  language processing}, pages 265--268. Association for Computational
  Linguistics, 1997.

\bibitem{loenneker2005narratological}
Birte L{\"o}nneker.
\newblock Narratological knowledge for natural language generation.
\newblock In {\em Proceedings of the 10th European Workshop on Natural Language
  Generation (ENLG-05)}, pages 91--100. Citeseer, 2005.

\bibitem{lukin2014automating}
Stephanie~M Lukin, James~O Ryan, and Marilyn~A Walker.
\newblock Automating direct speech variations in stories and games.
\newblock 2014.

\bibitem{MairesseWalker11}
Francois Mairesse and Marilyn~A. Walker.
\newblock Controlling user perceptions of linguistic style: Trainable
  generation of personality traits.
\newblock {\em Computational Linguistics}, 2011.

\bibitem{montfort2007generating}
Nick Montfort.
\newblock {\em Generating narrative variation in interactive fiction}.
\newblock University of Pennsylvania, 2007.

\bibitem{pennebaker2001linguistic}
James~W Pennebaker, Martha~E Francis, and Roger~J Booth.
\newblock Linguistic inquiry and word count: Liwc 2001.
\newblock {\em Mahway: Lawrence Erlbaum Associates}, 71:2001, 2001.

\bibitem{PennebakerSeagal99}
James~W Pennebaker and Janel~D Seagal.
\newblock Forming a story: The health benefits of narrative.
\newblock {\em Journal of clinical psychology}, 55(10):1243--1254, 1999.

\bibitem{Prince73}
Gerald Prince.
\newblock {\em A Grammar of Stories: An Introduction}.
\newblock Number~13. Walter de Gruyter, 1973.

\bibitem{Propp68}
Vladimir~Iakovlevich Propp.
\newblock {\em Morphology of the Folktale}, volume~9.
\newblock University of Texas Press, 1968.

\bibitem{rishes2013generating}
Elena Rishes, Stephanie Lukin, David~K. Elson, and Marilyn~A. Walker.
\newblock Generating dierent story tellings from semantic representations of
  narrative.
\newblock In {\em International Conference on Interactive Digital Storytelling,
  ICIDS'13}, 2013.

\bibitem{Shklovsky91}
Viktor Shklovsky.
\newblock {\em Theory of prose}.
\newblock Dalkey Archive Press, 1991.

\bibitem{SlaterRouner02}
Michael~D Slater and Donna Rouner.
\newblock Entertainment education and elaboration likelihood: Understanding the
  processing of narrative persuasion.
\newblock {\em Communication Theory}, 12(2):173--191, 2002.

\bibitem{Traumetal07}
D.~Traum, A.~Roque, A.~L.~P. Georgiou, J.~Gerten, B.~M.~S. Narayanan,
  S.~Robinson, and A.~Vaswani.
\newblock Hassan: A virtual human for tactical questioning.
\newblock In {\em Proceedings of SIGDial}, 2007.

\bibitem{Walkeretal11}
M.A. Walker, R.~Grant, J.~Sawyer, G.I. Lin, N.~Wardrip-Fruin, and M.~Buell.
\newblock Perceived or not perceived: Film character models for expressive nlg.
\newblock In {\em International Conference on Interactive Digital Storytelling,
  ICIDS'11}, 2011.

\end{thebibliography}

\end{document}